# A MAC-less Neural Inference Processor Supporting Compressed, Variable Precision Weights

Vincenzo Liguori     Ocean Logic Pty Ltd          enzo@ocean-logic.com

**Abstract**
**This paper introduces two architectures for the inference of convolutional neural networks (CNNs). Both architectures exploit weight sparsity and compression to reduce computational complexity and bandwidth. The first architecture uses multiply-accumulators (MACs) but avoids unnecessary multiplications by skipping zero weights. The second architecture exploits weight sparsity at the level of their bit representation by substituting resource-intensive MACs with much smaller Bit Layer Multiply Accumulators (BLMACs). The use of BLMACs also allows variable precision weights as variable size integers and even floating points. Some details of an implementation of the second architecture are given. Weight compression with arithmetic coding is also discussed as well as bandwidth implications. Finally, some implementation results for a pathfinder design and various technologies are presented.**

# 1. Introduction

Neural networks (NNs) and, particularly, convolutional neural networks (CNNs) are used in many applications. The high computational cost of their inference constitutes a challenge in low power and low cost environments such as embedded system and IoT.

This paper aims to address some of these limitations and greatly reduce the computational and bandwidth cost of CNN inference. The architectures introduced are based on the following characteristics:

- Efficient convolution in sparse CNNs with minimal on-chip memory feature map caching and external memory accesses.
- Weight compression to further reduce external memory bandwidth
- Elimination of multipliers for CNN inference, replaced by BLMACs while allowing variable precision weights.

Knowledge of the techniques discussed in section 4 in [1] is essential for understanding this paper.

# 2. Convolutions

This section introduces an architecture designed to process convolutions in sparse CNNs efficiently. The architecture constitutes the basis for subsequent more efficient architectures. The aim is to minimise on-chip memory for feature map caching requirements and external memory accesses.

## 2.1. Feature Maps

This paper focuses on the convolutional layers as they are at the most computationally intensive of CNNs.

In a convolutional layer, the input, a 3D set of values with size [X,Y,Z] (the input feature map), is mapped to another 3D set of values with size [$X_1,Y_1$, O] (the output feature map).  This paper refers to the elements of feature maps as "pixels".

Such mapping is performed by convolving a set of filters (also known as kernels) with the input feature map (with strides of $S_x$ pixels in the x direction and $S_y$ pixels in the y direction). The elements of the kernel W (also known as weights) form a tensor of size [K,K,Z,O] (with K usually an odd integer). Each pixel of the output feature map is:

$$pixout_{\frac{x}{S_x},\frac{y}{S_y},o} = \sum_{z=0}^{Z-1} \sum_{i=0}^{K-1} \sum_{j=0}^{K-1} pixin_{x+j-K/2, y+i-K/2, z} W_{j,i,z,o} + b_o \quad (1)$$

The values $b_o$ are "biases". Eq. 1 is valid for 0≤x<X, 0≤y<Y and x mod $S_x$ = y mod $S_y$ = 0 with mod being the arithmetic module. It follows that $X_1$=X/$S_x$ and $Y_1$=Y/$S_y$ (integer division). For convolutions performed at the edges of the input feature map, part of the kernel falls outside of said map. In this paper, the input feature map is assumed padded with zeros (padding with boundary pixels is another option). We also assume $S_x$=$S_y$=1. This is true for most modern CNNs, and, in any case, it should not be difficult to extrapolate when this is not the case.

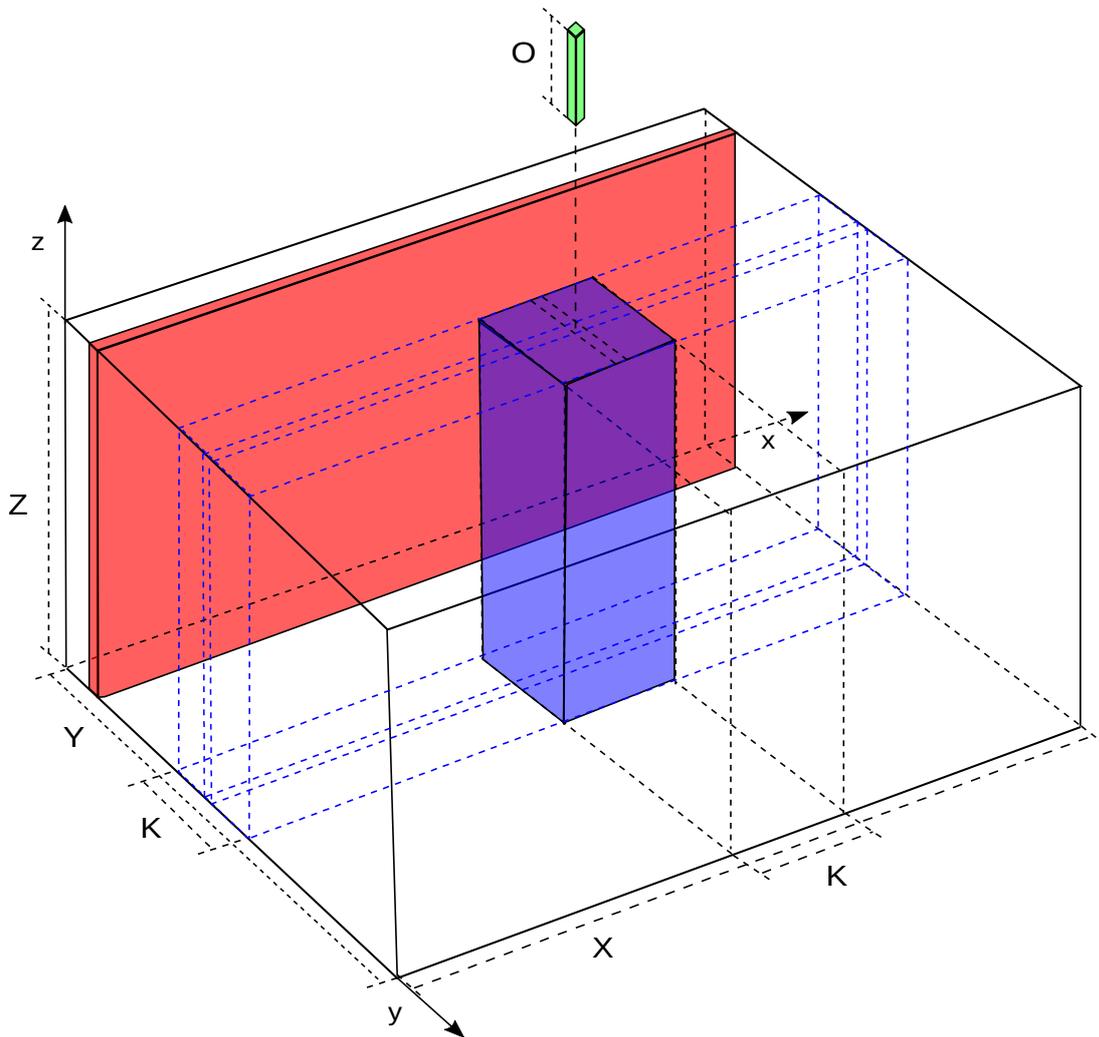

Figure 1 Input feature map.

Fig. 1 illustrates an input feature map of [X,Y,Z] pixels. In this paper, a "slice" (in red in the figure above) is all the pixels that have the same y coordinate. A "row" is all the pixels that have the same `y` and `z` coordinates. It follows that a slice in the input feature map consists of a Z rows stacked together. A column is all the pixels that have the same `x` and `y` coordinates. The volume of pixels in blue shows the pixels that are required to calculate a column of O pixels in the output feature map according to (1). It is an "active area" centred around a particular `x,y` point. Sliding the kernel along the x axis and calculating a column of O pixels of the output feature map at each valid point generates a slice of the output feature map. It is also worth noticing that only K adjacent slices of the input feature map are required at any one time to calculate a slice of the output feature map.

From (1) it is clear that calculating output feature maps is an embarrassingly parallel problem. There are many ways to perform such computation. The strategy chosen here aims to minimise the number of external memory accesses and the amount of pixel caching.

To satisfy this objective, the architecture loads slices from the input feature map in sequential order from $y=0$ to $y=Y-1$ from external memory. The slice sequential order is not arbitrary. CNNs are most common in image processing where an X*Y RGB image is an input feature map [X,Y,3]. Such images are normally acquired from raster device such as a CMOS sensor, and each sequentially incoming video line represents a slice of such an input feature map. This sequential access (both writing and reading) of the feature maps, one slice at the time, increases the efficiency of external memories such as DRAM.

Slices from the output feature map are also computed sequentially from slice 0 to Y-1 and eventually stored back to external memory. Note that computing a single output slice needs no more than K adjacent input slices. Indeed this is the minimum size of the on-chip cache (Slice Buffer). However, it is beneficial to cache K+1 slices so that the computation of the output slice (performed on K slices) and the loading of the next input slice from external memory can overlap.

Fig.2 shows the overlap. The Slice Buffer is, effectively, a shift register. Each time a new slice is loaded, the ones already present are shifted to the left. After enough slices (K/2+1) have been loaded, the Processing Elements (PEs) can start calculating one output slice. While the PEs are working, the next input slice can be loaded in the Kth slot, which allows the overlapping of computation and external memory accesses.

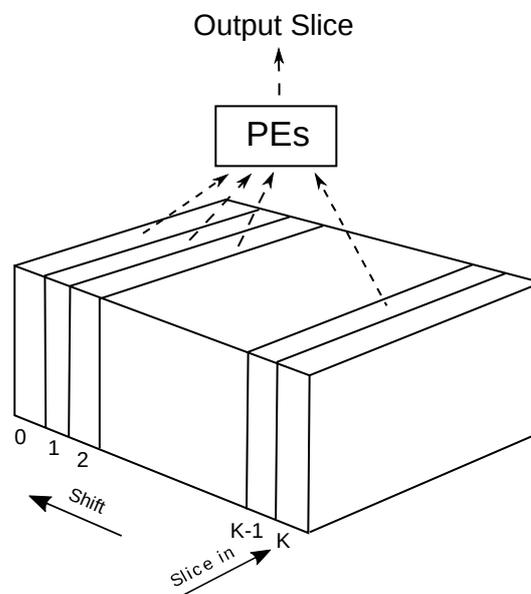

*Figure 2 Slice Buffer operations.*

Next, to address the problem of storing feature maps in on-chip RAM, consider these points:

- To compute an output feature map only requires storing K+1 slices (from 0 to K), regardless of the size of the input or output feature maps. Requiring so little storage makes it practical to use on-chip memory.

- Computing an output slice and loading the next input slice happen concurrently. As long as the time it takes to load one input slice is the same or less than the time to compute one output slice, the computation can proceed uninterrupted by memory accesses.

- Both input and output feature map slices are input and computed sequentially from slice 0 onward.

- The design is independent of the order of loading the pixels within a slice $y$ of the input feature map.

- The design is independent of the method and order of computing the pixels within an output feature map slice.

Once started, pixels from the input feature map incoming slice are always written to the Kth position in the Slice Buffer. Once a full slice has been input, the content of the buffer is shifted, and the pixels in slice buffer 1 are moved to slice buffer 0, the one in 2 to 1 and so on, with the pixels in Kth slice buffer moved to slice buffer K-1. Pixels in slice buffer 0 are no longer needed and are dropped. This shifting creates space in the Kth slice buffer for another incoming input feature map slice.

Shifting data is only conceptual. Actually shifting slices within the Slice Buffer would take too much time and power. Rather, it is only necessary to update pointers to the memory addresses of the $0^{th}$ and $K^{th}$ slice.

No computation of the output map slice starts until the first K/2+1 slices have been input. This is the same as the convolutions (1) being performed on the edge of the input feature map while the kernel falls partly outside said edge. Processing treats the non- existing slices as zeros. A similar situation occurs when processing the last slices of the input feature map. Therefore, the PE(s) need to keep track of when the kernel falls off the edges of the input feature map and pad the input with zeros.

Only K+1 slices need to be stored, which is normally about two orders of magnitude smaller than the size of a feature map (Y slices). Therefore, it is practical to implement the pixel cache in on-chip memory.

Note that accesses to the external memory are sequential, allowing almost uninterrupted bursts from and to DRAM.

## 2.2. Processing Elements – Architecture I

The PEs are responsible for calculating an output slice from the cached K input slices. Ideally, they should also:

- efficiently calculate the output slice when the weights are sparse (i.e. skipping all the zero weights and thus avoiding multiplications by zero)
- use a weight only once while calculating a given output slice (i.e. avoid re-loading/re-obtaining the same weight again while calculating the same output slice)

The latter is allows decompressing non-zero weights from a sequential stream.

Eq. (1) shows that, to calculate a pixel of the output feature map, the input feature map pixels are multiplied by the weights, and the results are accumulated.

This architecture uses an array of MACs, one for each of the X pixels in a row in order to calculate a row of pixels of the output slice. For each output $o$ of (1), we initialise each MAC accumulator with zero and proceed to multiply and add all the $pixin_{x+j-K/2,y+i-K/2,z}$ by $W_{j,i,z,o}$ products simultaneously for all the X pixels in a row. This calculates Eq. 1 minus the bias $b_o$.

Fig.3 shows architecture I. It calculates a row of X pixels of the output feature map slice y for a given index o using (1). Each of the X MAC registers contains the pixel $pout_{x,y,o}$.

At its input, the MAC requires the value of the weights $W_{j,i,z,o}$ as well as its indexes `i,j,z` (`o` is assumed fixed while calculating the `o`$^{th}$ row of pixels).

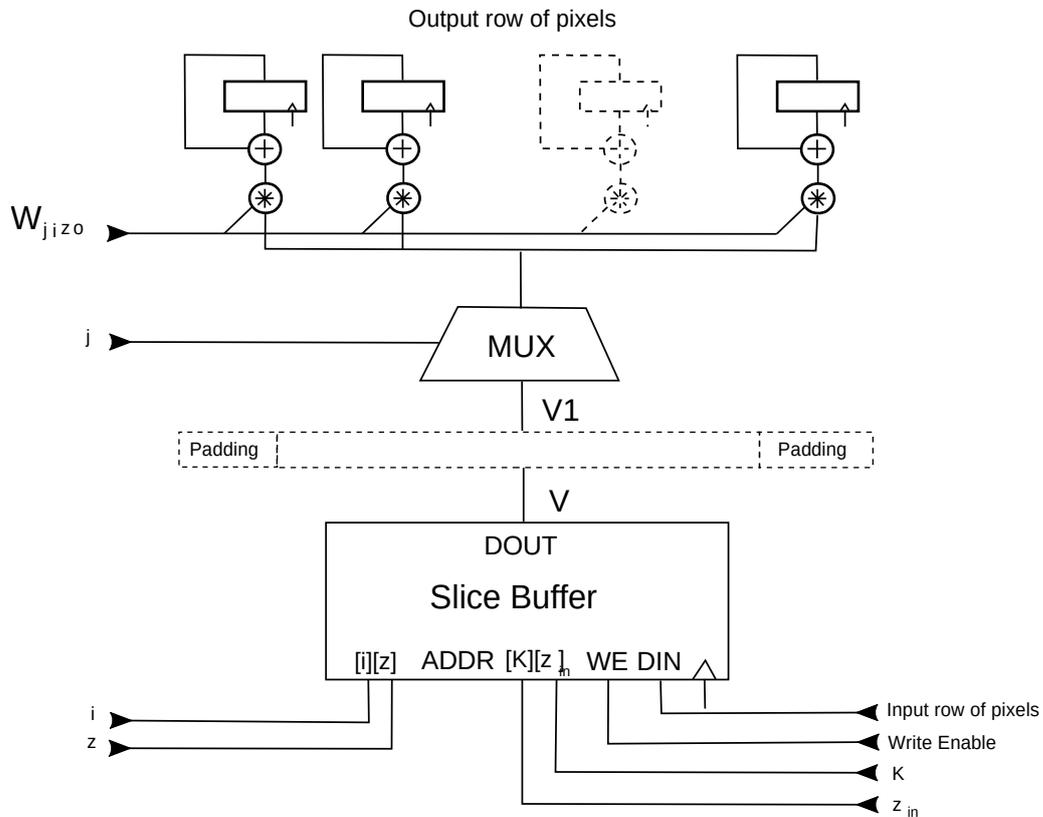

*Figure 3 PE architecture I.*

The slice buffer arranged as a 2D array Slice_Buffer[K+1][Z] of words X pixels wide and is implemented as shown in Fig2. The memory contains K+1 slices of a feature map, each of which is formed by Z rows of X pixels (see also Fig.1).

Conceptually, the SliceBuffer is a dual-port memory. Incoming rows of pixels from external memory are written to Slice_Buffer[K][$z_{in}$], and rows of pixels are read from SliceBuffer[i][z]. In practice, there are various ways to implement the SliceBuffer that do not require the area and timing cost of a true dual-port memory. For example, by assigning a bank of RAM for each of the K+1 slices, since a write and a read operation never happen in the same slice, it is possible to have dual-port functionality with only single-port RAMs.

For simplicity, the diagram above omits the mechanism that allows the shifting of the slices in the SliceBuffer at the end of each output slice calculation. Only simple pointer arithmetic on a small number of bits (in most cases K=3, 2 bits only) is needed.

The calculation begins by initializing the MACs' accumulators to zero for the $o^{th}$ row of pixels, of a given slice y.

Then, for each weight $W_{j,i,z,o}$ and its indexes j, I, and z, word V is retrieved from SliceBuffer[i][z].V is padded, both left and right, by K/2 zero pixels, forming a larger word V1.

The architecture keeps track of the slice number y, and uses that to determine when the kernel falls off the input feature map, and the next row of pixels is assumed to be zeros.

The mux selects X pixels V1[x+j] (0≤x<X), simultaneously multiplies them by the same weight $W_{j,i,z,o}$, and accumulates the result in the MAC array. By pipelining the design, this can be done at the rate of a row of pixels per clock cycle. After processing all the $W_{j,i,z,o}$ (for a particular index o), the MAC accumulators contain the $o^{th}$ row of pixels of a slice of the output feature map (minus the bias $b_o$) and can be output. An End Of Run (EOR) signals the last weight. The process repeats for a different index o.

Note that this architecture, with a given, fixed, index o, can use the weights $W_{j,i,z,o}$ one at a time and in any order. Being order-independent allows skipping from zero weights in sparse vectors. Using weights one at a time is helpful because that is how weights are decompressed.

## 2.3. Weights Encoding and Compression

The architecture described above processes one weight per clock cycle even with weights in no particular order (for a given output slice row o). This architecture works well with sparse weights as it allows jumping from one non-zero weight to another, provided that there is no clock cycle penalty for the logic to skip zero weights.

Each non-zero weight can be encoded as a (ZRUN,W) pair using a run-length scheme. Here ZRUN represents the number of zero weights preceding the weight W. For a run-length scheme to work, the tensor W[K][K][Z][O] needs to be flattened into a vector.

Once flattened, the weights can be encoded as (ZRUN,W) pairs. The pair (0,0) can be the special symbol EOR, which indicates the end of a run-length sequence. EOR indicates that all the remaining elements are zero and, thus, do not contribute to the calculation. EOR need not be explicitly coded if the last element of a flattened tensor is non-zero (for a given o).

The (ZRUN,W) pairs can be further compressed with an entropy encoder such as Huffman or arithmetic encoding. This reduces the amount of storage needed for each o from 0 to O-1. Note that both storage and retrieval of the compressed weights to/from external memory are in sequential locations, which is much more efficient than random access in DRAM.

The process can be reversed when the weights are needed by the PEs to perform the convolutions. The compressed weights are fetched from the external memory with the additional benefit of the reduced bandwidth due to compression. An entropy decoder reconstructs the sequence of (ZRUN,W) pairs, which is expanded to recreate the sequence of non-zero weights.

This process gives architecture I a stream of non-zero weights and associated `i,j,z` and EOR, at a rate of up to one per clock cycle.

Fig.4 shows the weigh compression and decompression processes described above. Compression happens off-line, and decompression happens in hardware at inference time.

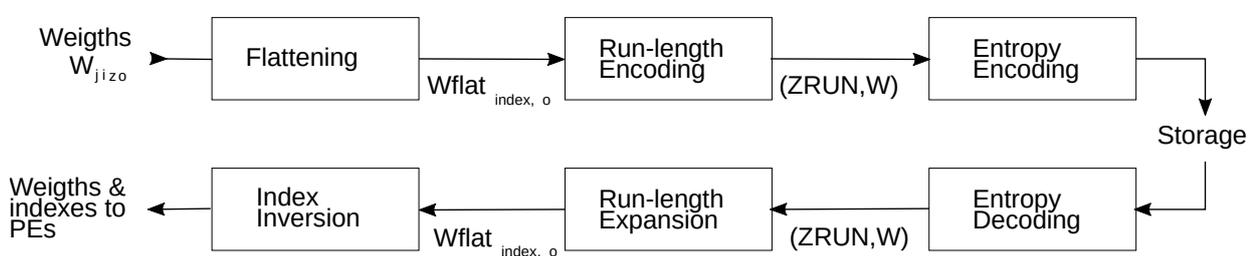

*Figure 4 Weigth run-length encoding and compression.*

# 3. Architecture II

Architecture II (Fig.5) is derived from architecture I by substituting the add/sub accumulators with BLMACs. The main changes are :

- The weights controlling the add/sub behaviour of the BLMACs are just one bit.

- The downstep counter lcnt counts the bit layers. The architecture calculates a row of pixels from the output slice using the bit layer method previously described.

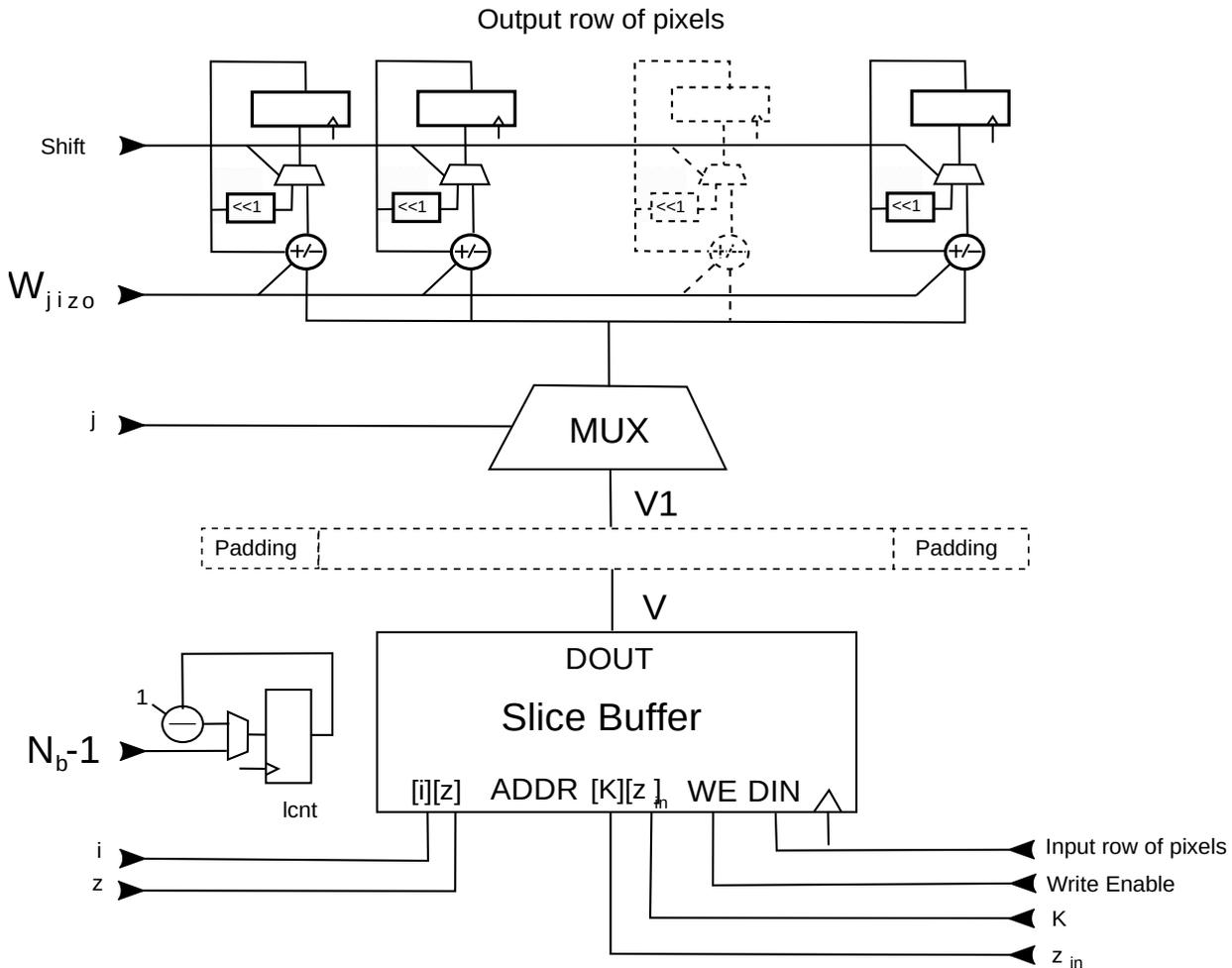

*Figure 5 PE architecture II.*

At the start of a row o calculation, the counter `lcnt` is loaded with $N_b$-1, where $N_b$ is the number of bit layers. The accumulators are also initialized to zeros. A stream of non-zero weights is input with its associated `i,j,z` parameters. The row of pixels retrieved from indexes `i,j,z` is added or subtracted from the accumulators according to the sign of the ±1 weight. If an EOR code is received and `lcnt` is non-zero, all of the accumulators are multiplied by 2 (using the shift input), which concludes a bit layer. `lcnt` is decremented. Otherwise, if `lcnt` was zero (after Nb EORs), row o of the output slice (minus the $b_o$ bias) is ready.

## 3.1. BLMAC Floating Point Performance

BLMAC integer performance is discussed in [1]. A BLMAC can perform dot products for CNNs efficiently because, as shown in [2], weights tend to have a Laplacian/Gaussian distribution. This efficiency also occurs when weights are quantized to integers.

A BLMAC can also, without modifications, perform dot products with floating-point weights. Floating point numbers are just large integers with, at most, the fractional part of non-zero bits (up to a power of 2 that can be taken into account when scaling). Therefore, multiplying by a floating-point number with a BLMAC is equivalent to multiplying by an integer of the size of the fractional part.

Note that, even if the weights tend to have a Laplacian/Gaussian distribution, the fractional part of their floating-point representation probably does not.

Assuming that the fractional part of a floating-point number has a uniform distribution, the table below shows the average number of cycles that a BLMAC would take to multiply by a weight and accumulate the result for various floating-point representations. This information is obtained from Tab. 5 in [1].

Using a BLMAC for floating-point numbers uses more cycles than a single cycle floating-point MAC. However, a BLMAC achieves the same result with, essentially, just the logic of an integer add/sub accumulator.

| Type | Sign | Exponent | Fractional part | BLMAC cycles |
|---|---|---|---|---|
| Half-precision | 1 | 5 | 10 | ~3.77 |
| Bfloat16 | 1 | 8 | 7 | ~2.77 |
| TensorFloat-32 | 1 | 8 | 10 | ~3.77 |
| Single-precision | 1 | 8 | 23 | ~8.11 |

Table 1: Average number of cycles for a fp multiply/accumulate operation with a BLMAC.

Although BLMACs can support floating-point weights, integer weights are generally better because they tend to have a Laplacian/Gaussian distribution whereas the fractional part of floating-point weights is essentially random.

# 4. Hardware Implementation

A hardware implementation of a neural inference processor based on architecture II was realised in synthetisable RTL. This pathfinder implementation aims at supporting features in Tiny Yolo v3 [3], although it is generalisable. Tests were performed on a 416x416 input image using the Darknet software [4] as reference. The TinyYolo v3 weights are pyramid vector quantized [5] as described in [1] and in section 2.3. Remember that the architecture performance depends on the amount of quantization. Some comparison videos for the given quantization are visible on Youtube [6].

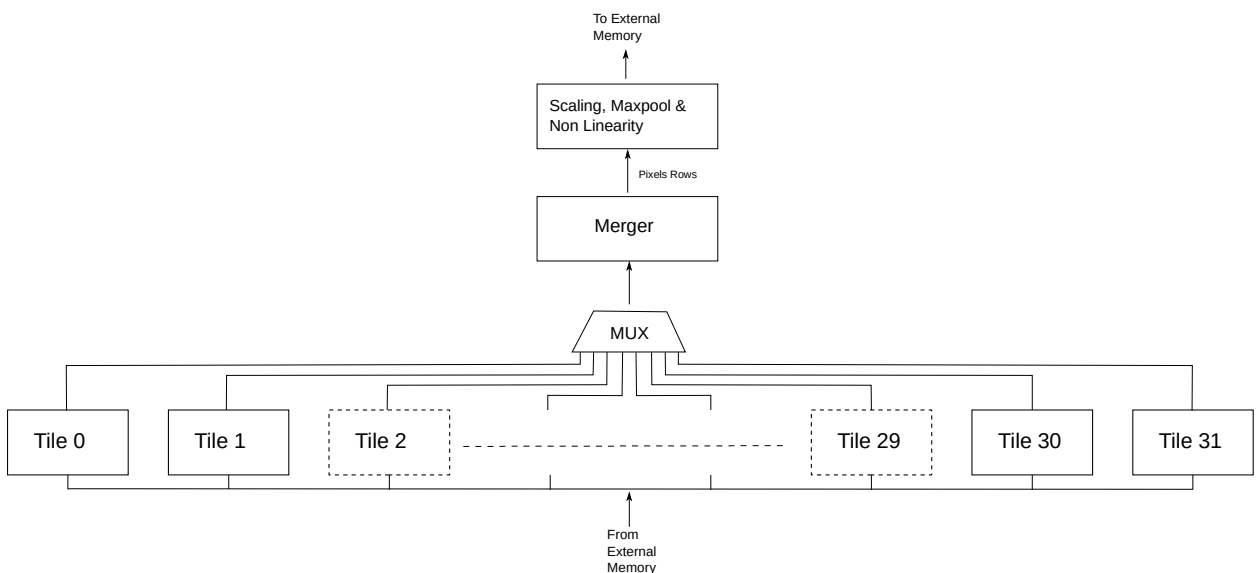

Figure 6 Inference processor architecture.

## 4.1. Implementation Details

The implementation consists of an array of 416 20 bits BLMACs, divided into 32 tiles of 13 BLMACs each, supporting 3x3 (K=3) and 1x1 (K=1) kernels. The Slice Buffer in each tile is a single-port 128x104 RAM, which is enough to store 4 slices (K+1 slices, as explained in section 2.2). Input and output pixels are 8 bits each.

Each tile has a weight decompression engine and cache. The cache is a sequentially-addressed memory. There is no complex management logic. Fig7. shows the tile architecture. The PE consists of 13 BLMACs, as per architecture II (Fig.5). The decompression engine uses arithmetic coding, and is capable of extracting one multi-bit symbol plus multiple uncompressed (bypass) bits in a single clock cycle. This produces a run-length code in each clock cycle, which keeps the BLMAC arrays operating continuously.

The arithmetic coding probability model is fully programmable. For simplicity, just two contexts (for zero and non-zero bit layers) are used in the pathfinder architecture. A separate context for each bit layer should give better compression results. The decompression engine in Fig.7 is very small (~250 LUT, including the programmable probability model) with very shallow logic that provides high clock speed (up to ~700 MHz in isolation in Xilinx Ultrascale+).

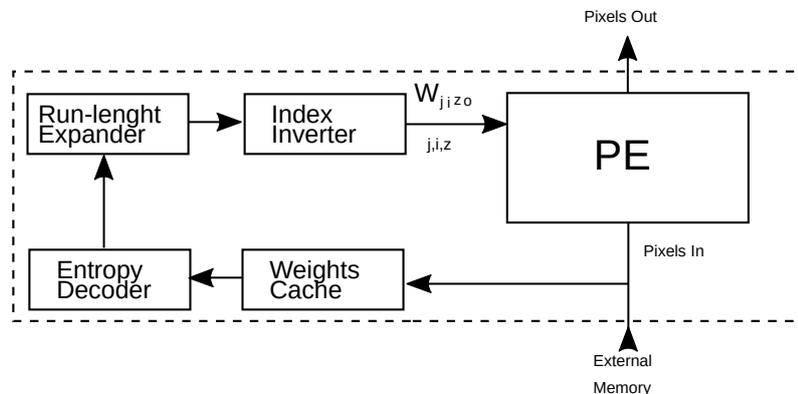

*Figure 7 Tile architecture.*

In the case of Tiny Yolo v3, as implemented in Darknet software, feature maps go from [416,416,3] to [13,13,1024]. This dimension changing is similar to other CNNs. The architecture II in Fig.5 is a fixed size row of BLMACs. More flexibility is needed to deal with different row sizes within different layers. Otherwise, some BLMACs are idle for narrow slices. Therefore, the array can be configured for the 32 tiles as a single array of 416 BLMACs working on a 416 pixels line, as two arrays of 208 BLMACs working together on a 208 pixels line, etc., all the way to 32 arrays of 13 BLMACs working in parallel on the same line of pixels.

The partial results from the tiles in different arrangements are then merged into a single line of pixels by the merger unit and funnelled to the scale/maxpool unit. The resulting pixels are passed through the Leaky ReLu non-linearity and scaled to 8 bits each. If required, the scaling unit can also perform a maxpool operation without accessing external memory. The processor then outputs a slice of pixels, line by line.

## 4.2. Operations

The operation of the processor in this pathfinder implementation is simple. For each layer, first, the compressed weights and their probability models are loaded into the weight caches in the tiles. Depending on their size, weights for more than one layer can be loaded. Next, the first K-1 slices are loaded in the Slice Buffer (see section 2.2). Computation can then start producing output slices. After each output slice is computed, a new slice is loaded in the Slice Buffer.Processing continues until the full output feature map is complete.

## 4.3. Performance

The first 5 columns of Tab. 2 contain layer information, including the input and output feature maps and the kernel applied. Each kernel requires a number of MAC operations equal to the product of its dimensions. For instance, for layer 0, the kernel is 3*3*3*16 = 432, which is the number of MAC operations required to apply the kernel and produce a column of pixels of an output feature map.

The next column shows the number of clock cycles taken by the pathfinder implementation described above to output a slice of pixels. These numbers were obtained by cycle-accurate RTL simulations. These included all of the steps described above, including convolutions, scaling, non-linearity and maxpooling. In case of maxpooling, only one slice every two is output (four adjacent pixels are merged into one), but each slice still needs to be calculated. Multiplying the number of cycles per output slice by the number of slices gives the number of cycles taken to complete an output feature map. These numbers establish a lower bound for the clock necessary to process a given number of frames per second. For example, for 30 frames/s, all the number of cycles per map require 7,909,915 cycles. That, multiplied by 30, gives ~ 237 MHz.

This clock frequency represents a lower bound because the figures in Tab. 2 do not include the loading of the weight caches or pixel loading and storing. These factors need to be taken into account for a practical implementation, as is calculated below and shown in Tab. 3.

| # | Layer | Kernel | Input | Output | MACs/Kernel | Cycles/Slice | Cycles/Map | Cycles/Kernel |
|---|---|---|---|---|---|---|---|---|
| 0 | Conv | 3x3x3x16 | 416x416x3 | 416x416x16 | 432 | 723 | 300,768 | 723 |
| 1 | Maxpool | 2x2/2 | 416x416x16 | 208x280x16 | | | | |
| 2 | Conv | 3x3x16x32 | 208x280x16 | 208x280x32 | 4,608 | 1,937 | 402,896 | 3,874 |
| 3 | Maxpool | 2x2/2 | 208x280x32 | 104x104x32 | | | | |
| 4 | Conv | 3x3x32x64 | 104x104x32 | 104x104x64 | 18,432 | 4,172 | 433,888 | 16,688 |
| 5 | Maxpool | 2x2/2 | 104x104x64 | 52x52x64 | | | | |
| 6 | Conv | 3x3x64x128 | 52x52x64 | 52x52x128 | 73,728 | 8,721 | 453,492 | 69,768 |
| 7 | Maxpool | 2x2/2 | 52x52x128 | 26x26x128 | | | | |
| 8 | Conv | 3x3x128x256 | 26x26x128 | 26x26x256 | 294,912 | 17,978 | 467,428 | 287,648 |
| 9 | Maxpool | 2x2/2 | 26x26x256 | 13x13x256 | | | | |
| 10 | Conv | 3x3x256x512 | 13x13x256 | 13x13x512 | 1,179,648 | 37,142 | 482,846 | 1,188,544 |
| 11 | Maxpool | 2x2/1 | 13x13x512 | 13x13x512 | | | | |
| 12 | Conv | 3x3x512x1024 | 13x13x512 | 13x13x1024 | 4,718,592 | 171,762 | 2,232,906 | 5,496,384 |
| 13 | Conv | 1x1x1024x256 | 13x13x1024 | 13x13x256 | 262,144 | 9,013 | 117,169 | 288,416 |
| 14 | Conv | 3x3x256x512 | 13x13x256 | 13x13x512 | 1,179,648 | 36,538 | 474,994 | 1,169,216 |
| 15 | Conv | 1x1x512x255 | 13x13x512 | 13x13x255 | 130,560 | 16,596 | 215,748 | 531,072 |
| 18 | Conv | 1x1x256x128 | 13x13x256 | 13x13x128 | 32,768 | 4,257 | 55,341 | 136,224 |
| 21 | Conv | 3x3x384x256 | 26x26x384 | 26x26x256 | 884,736 | 52,111 | 1,354,886 | 833,776 |
| 22 | Conv | 1x1x256x255 | 26x26x256 | 26x26x255 | 65,280 | 12,515 | 325,390 | 200,240 |
| | | | Cycles per frame | | | | 7,909,915 | |

Table 2: Number of clock cycles for each output slice.

The last column of Tab.2 shows the number of cycles a single BLMAC in the pathfinder implementation actually needs to apply a kernel and generate a column of output pixels. This is generally 10-15% worse than the theoretical numbers shown in Tab 7 in [1] (where $N_3$ is the number of non zero trits in each kernel). The best case is layer 12, which is only ~5% worse.

The worst cases are layers 0, 15, 18 and 22. In the case of layer 0, this is because the scale and maxpooling unit only uses 7 multipliers and it is, therefore, slower than the BLMAC array at processing a slice. Though the operations of these two units largely overlap, in the case of layer 0 the scale and maxpooling unit ultimately stalls the BLMAC array. In the case of layers 15, 18 and 22, the bottleneck is ultimately the multiplexing system that connects the BLMAC array to the scale and maxpooling unit (i.e. the mux cannot clear a a pixel line as fast as the BLMACs compute it).

Note that, for computationally demanding layers, one BLMAC performs roughly as well as one MAC. For example, applying the layer 12 kernel requires 4,718,592 MAC operations.The BLMAC takes 5,496,384 cycles. Assuming that each MAC operation takes one clock cycle, that is ~16% more. However, a

BLMAC is about one order of magnitude smaller than a MAC and can run at higher clock speed in many technologies.

## 4.4. Bandwidth Requirements

Tab. 3 lists, for each layer, the number of bytes that need to be loaded and stored. This includes the input and output of the feature maps (starting with a 416x416 image) and the compressed weights.

The size of the input and output feature maps is the product of its three dimensions. So, for example, the input colour image is 416x416x3 = 519,168 bytes.

The totals at the bottom of the table give the amount of data transferred by the processor from and to the external memory for each image processed. The totals are useful for estimating the bandwidth required by the processor for a given frame rate.

| # | Layer | Input | Input Map | Output | Output Map | Compressed Weights |
|---|---|---|---|---|---|---|
| 0 | Conv | 416x416x3 | 519,168 | 208x280x16 | 692,224 | 288 |
| 1 | Maxpool | | | | | |
| 2 | Conv | 208x280x16 | 692,224 | 104x104x32 | 346,112 | 1,788 |
| 3 | Maxpool | | | | | |
| 4 | Conv | 104x104x32 | 346,112 | 52x52x64 | 173,056 | 7,506 |
| 5 | Maxpool | | | | | |
| 6 | Conv | 52x52x64 | 173,056 | 26x26x128 | 85,258 | 31,372 |
| 7 | Maxpool | | | | | |
| 8 | Conv | 26x26x128 | 85,258 | 13x13x256 | 43,264 | 131,476 |
| 9 | Maxpool | | | | | |
| 10 | Conv | 13x13x256 | 43,264 | 13x13x512 | 85,258 | 526,850 |
| 11 | Maxpool | | | | | |
| 12 | Conv | 13x13x512 | 85,258 | 13x13x1024 | 173,056 | 1,811,304 |
| 13 | Conv | 13x13x1024 | 173,056 | 13x13x256 | 43,264 | 123,982 |
| 14 | Conv | 13x13x256 | 43,264 | 13x13x512 | 85,258 | 511,406 |
| 15 | Conv | 13x13x512 | 85,258 | 13x13x255 | 43,095 | 63,768 |
| 18 | Conv | 13x13x256 | 43,264 | 13x13x128 | 43,264 | 16,460 |
| 21 | Conv | 26x26x384 | 259,584 | 26x26x256 | 173,056 | 369,792 |
| 22 | Conv | 26x26x256 | 173,056 | 26x26x255 | 172,380 | 30,964 |
| | Total | | 2,721,822 | | 2,158,545 | 3,626,956 |

Table 3: Number of bytes for input and output maps as well as the compressed weights.

As mentioned above, these transfers have a significant effect on overall performance. In the case of the Slice Buffer, as explained in section 2.1, loading pixels can happen concurrently to the output slice calculation, effectively "hiding" the load time. As mentioned above, this requires implementing the Slice Buffer either with multiple RAM banks or with dual-port RAM.

Note that all of the external memory accesses are sequential. Pixel lines and compressed weights are read and stored sequentially. Sequential accesses can be handled with long bursts to the external memory, which is particularly advantageous for DRAM. The accesses are also predictable, allowing for predictable scheduling.

Furthermore, note that processing different sized feature maps does not change the memory bandwidth related to the weights. Specifically, for larger feature maps, the weight bandwidth is proportionally smaller.

Finally, the size of the weight cache affects performance. Specifically, if a layer's compressed weights do not fit into the cache, multiple passes are needed. This is because the weight tensor needs to be split so that each portion, once compressed, fits in the available weight cache. The tensor is split along the o coordinate (see section 2.1). We now need multiple passes where, at each pass, a portion of the compressed weights is loaded into the weight cache with the processor then producing a number of output lines. This is repeated multiple times until a full output slice is produced.

We can see that these operations do not affect the weight cache bandwidth as the total weight size remains the same. Same for the output bandwidth, as the output remains the same. There is, however, an effect on the input bandwidth as the same input feature map is loaded into the Slice Buffer at each pass.

Note that this does not necessarily affect the computation speed, as long as the loading of each slice can be "hidden" during the computation, as shown in section 2.1. It does, however, affect the bandwidth required as well as the power, due to the additional accesses to the external memory.

## 4.5. Resource Utilization

Tab. 4 shows the number of resources required and the maximum clock frequency achieved for the pathfinder implementation in various technologies. The design is a single, rising edge only clock, fully synchronous design with no transparent latches.

| Technology | Logic | Memory | DSPs | Frequency | TOPs |
|---|---|---|---|---|---|
| Xilinx Artix-7 | ~33 K LUTs | 3 RAMB36 | 7 | ~240.3 MHz | ~0.169 |
| Xilinx Artix-7 | ~26 K LUTs | 35 RAMB36 + 32 RAMB18 | 7 | ~246.5 MHz | ~0.174 |
| Xilinx Ultrascale+ | ~33 K LUTs | 3 RAMB36 | 7 | ~543.7 MHz | ~0.383 |
| Xilinx Ultrascale+ | ~26 K LUTs | 35 RAMB36 + 32 RAMB18 | 7 | ~502.3 MHz | ~0.354 |
| Intel Cyclone V | ~26.7 K ALMs | 105 M10K | 7 | ~176.1 MHz | ~0.124 |
| Intel Cyclone V | ~34.3 K ALMs | 9 M10K | 7 | ~150.2 MHz | ~0.106 |
| Intel Stratix V | ~25.6 K ALMs | 101 M20K | 7 | ~308.2 MHz | ~0.217 |
| Intel Stratix V | ~33 K ALMs | 5 M20K | 7 | ~276.6 MHz | ~0.195 |
| Intel Arria10 | ~29.4 K ALMs | 101 M20K | 7 | ~280.7 MHz | ~0.198 |
| Intel Arria10 | ~37.7 K ALMs | 5 M20K | 7 | ~273.3 MHz | ~0.192 |
| ASIC 7 nm | ~371 Kgates | ~561 Kbits | - | ~2.0 GHz | ~1.41 |

Table 4: Resource utilisation for the described architecture.

For each of the FPGA families, the Slice Buffer is implemented as either distributed or block RAM. Note that, for Xilinx, two 24 bit BLMACs could be implemented in a single DSP48 in SIMD mode, allowing for extra flexibility and resources.

For ASIC, 1 gate = 1 NAND2 gate. All the memories are single-port, except for ~56 Kbits.

The figures above do not include the memories required for the weight caches. For the sake of obtaining a result for synthesis for FPGAs, ~2 Mbits of block RAMs were used (i.e. 64 RAMB36, two per tile).

According the Darknet web page [4], TinyYolo v3, on a 416x416 image, requires 5.56 billions of floating point operations. From this, and knowing the number of clock cycles taken by this design, we can extrapolate the number of operations per second for a given clock. So, as per Tab.2, an image takes ~7.9e+6 cycles and that gives us 5.56e+9/7.9e+6 = ~704 operations/clock. Scaling this by the clock frequency, we obtain the figures in the last column.

This pathfinder architecture is useful, even in low-end FPGAs. For example, the Darknet software [4], by default, processes 416x416 images. If input images are of different sizes, they are scaled by a necessary factor to fit for processing and then scaled back by the inverse factor after processing. Thus, for a

640x480 image, the processing is on 416x320 pixels. Using Tab.2 to re-calculate the lower bound for the clock for a 416x320 image at 30 fps, we require ~182.5 MHz. At such low clock speed requirements, even with realistic external memory accesses and weight caches sizes, low end FPGAs can process useful images at full frame rate (24-30 fps).

# 5. Conclusion

This paper has introduced an architecture for neural inference acceleration based on BLMACs. Some characteristics include:

- Low use of resources and shallow, fast logic
- Low memory and bandwidth requirements for both pixels and weights
- Support for variable precision, compressed weights

Results for an actual implementation for a pathfinder architecture have been presented.